\documentclass[final]{cvpr}

\usepackage{times}
\usepackage{epsfig}
\usepackage{graphicx}
\usepackage{amsmath}
\usepackage{amssymb}
\usepackage[ruled, lined, linesnumbered, commentsnumbered, longend]{algorithm2e}
\usepackage{siunitx}
\usepackage{booktabs}
\usepackage{tabto}

\usepackage[pagebackref=true,breaklinks=true,colorlinks,bookmarks=false]{hyperref}

\def\cvprPaperID{59} % *** Enter the CVPR Paper ID here
\def\confYear{CVPR 2021}

\begin{document}

%%%%%%%%% TITLE
% \title{\LaTeX\ Author Guidelines for CVPR Proceedings}
\title{Semi-Supervised Classification and Segmentation on High Resolution Aerial Images}

\author{Sahil Khose$^{*}$  \quad Abhiraj Tiwari$^{*}$  \quad Ankita Ghosh$^{*}$ \\
Manipal Institute of Technology, Manipal \\
{\tt\small \{sahil.khose, abhiraj.tiwari1, ankita.ghosh1\}@learner.manipal.edu}


}

\maketitle
{\let\thefootnote\relax\footnote{\begin{flushleft}
$^*$Authors have contributed equally to this work and share first authorship
\end{flushleft}}}

%%%%%%%%% ABSTRACT
\begin{abstract}
% Visual scene comprehension is a key requirement for decision making in any computer vision system. 
FloodNet is a high-resolution image dataset acquired by a small UAV platform, DJI Mavic Pro quadcopters, after Hurricane Harvey. The dataset presents a unique challenge of advancing the damage assessment process for post-disaster scenarios using unlabeled and limited labeled dataset. We propose a solution to address their classification and semantic segmentation challenge. We approach this problem by generating pseudo labels for both classification and segmentation during training and slowly incrementing the amount by which the pseudo label loss affects the final loss. Using this semi-supervised method of training helped us improve our baseline supervised loss by a huge margin for classification, allowing the model to generalize and perform better on the validation and test splits of the dataset. In this paper, we compare and contrast the various methods and models for image classification and semantic segmentation on the FloodNet dataset.
\end{abstract}

%%%%%%%%% BODY TEXT
\section{Introduction}
The frequency and severity of natural disasters threaten human health, infrastructure and natural systems. It is extremely crucial to have accurate, timely and understandable information to improve our disaster management systems. Rapid data collection from remote areas can be easily facilitated using small unmanned aerial systems which provide high-resolution images. Visual scene understanding of these collected images is vital for quick response and large scale recovery post-natural disaster. Classification and segmentation tasks are fitting in such situations as they can provide scene information to help the task force make decisions. 

One of the major challenges with generating a vision dataset is the cost of labeling the data, especially for semantic segmentation. This often leads to labels only for a small percentage of the data which gives rise to the need for semi-supervised methods that can produce results that are at par with supervised methods. Another challenge that we face apart from the lack of labeled dataset is the heavy class imbalance. Lack of labeled data coupled with class imbalance makes it a very challenging task to solve. Through our approach, we try to tackle these problems and produce creditable results.

Our contribution in this paper is two folds: semi-supervised classification and semi-supervised semantic segmentation. We discuss the existing literature in these fields based on which we have crafted our approach in Section \ref{section:related_works}. The subsequent sections throw light upon the pipeline of classification \ref{section:classification} and segmentation \ref{section:segmentation} respectively. Our paper describes the preprocessing and augmentation techniques used on the dataset, the methodology followed, the models and hyperparameters we have experimented with and the results obtained. Finally, we conclude by summarizing our results and discussing the scope of improvement in future \ref{section:conclusion}.

%-------------------------------------------------------------------------
\section{Related Works}
\label{section:related_works}

\textbf{Supervised Classification:}
Classification has been one of the earliest undertakings in the deep learning domain. Over the years, several architectures and methods have emerged which have leveraged extensive datasets like ImageNet \cite{imagenet} to produce state of the art results using supervised learning. The architectures that our paper makes use of are ResNet \cite{resnet} and EfficientNet \cite{efficientnet}.  

ResNet proposes residual connection architecture which makes it feasible to train networks with a large number of layers without escalating the training error percentage. Using the technique of skip connections, it resolves the issue of vanishing gradient.

EfficientNet proposes a simple but highly effective scaling method that can be used to scale up any model architecture to any target resource constraints while maintaining model efficiency. They observed the effects of model scaling and identified that carefully balancing network depth, width and resolution can lead to better performance.

\textbf{Supervised Semantic Segmentation:} Segmentation aids in extracting the maximum amount of information from an image. Instance segmentation treats every occurrence of a class as a unique object. Semantic segmentation is more coherent and associates every pixel of an image with a class label. Deep learning models like UNet\cite{unet}, PSPNet\cite{pspnet} and DeepLabV3+\cite{deeplabv3+} have provided exceptional results for this task. 

The architecture of UNet is divided into two parts: contracting path and expansive path. The contracting path follows the generic framework of a convolutional network while the expansive path undergoes deconvolution to reconstruct the segmented image.

PSPNet exploits the capability of global context information using different region-based context aggregation by introducing a pyramid pooling module along with the proposed pyramid scene parsing.

DeepLabV3+ is a refinement of DeepLabV3 which uses atrous convolution. Atrous convolution is a powerful tool to explicitly adjust the filter's field-of-view as well as control the resolution of feature responses computed by Deep Convolution Neural Network.

\textbf{Semi-supervised Approach:}
Pseudo-Label \cite{sudo_lbl} proposes a simple semi-supervised learning approach. The idea is to train the neural network in a supervised fashion with both labeled and unlabeled data simultaneously. For unlabeled data, pseudo labels are generated by selecting the class which has maximum predicted probability. This is in effect equivalent to Entropy Regularization \cite{entropy_reg}. It favors a low-density separation between classes, a commonly assumed prior for semi-supervised learning.

\section{Classification}
\label{section:classification}
The dataset has 2343 images of dimensions $3000\times4000\times3$ divided into training (1445), validation (450) and test (448) splits. Out of the 1445 training images 398 are labeled and 1047 are unlabeled. The labels for the classification task are Flooded and Non-Flooded.
In this section we describe our approach for classifying the FloodNet dataset \cite{floodnet} into 2 classes, Flooded and Non-Flooded.

\subsection{Data and Preprocessing}
In the given dataset 398 sample images were labeled out of which 51 samples were flooded and 347 were non-flooded. The large class imbalance prevents the model from achieving a good F1 score while training with the labeled dataset. To prevent this we used a weighted sampling strategy while loading the data in the model as inspired from \cite{r2021bert}. Both the classes were sampled equally during batch generation.

The labeled dataset was heavily augmented to get more images for training the model under supervision. The image samples were randomly cropped, shifted, resized and flipped along the horizontal and vertical axes.

We downsized the image from $3000\times4000$ to $300\times400$ dimensions to strike a balance between processing efficiency gained by the lower dimensional images and information retrieval of the high-resolution images. 

\subsection{Methodology}
ResNet18  with a binary classification head was used for semi-supervised training on the dataset. The model was trained for $E$ epochs out of which only the labeled samples were used for $E^{\alpha}_{i}$ epochs after which pseudo labels were used to further train the model. $\alpha$ has an initial value of $\alpha_{i}$ that increases up to $\alpha_{f}$ from epoch $E^{\alpha}_{i}$ to $E^{\alpha}_{f}$ as described in Algorithm \ref{alg:semi_cls}.

A modified form of Binary Cross-Entropy (BCE) was used as the loss function as shown in line 10 in Algorithm \ref{alg:semi_cls} where $l$ is the label of a sample, $\hat{l}$ is the predicted class for labeled sample and $u_{epoch}$ is the predicted class for an unlabeled sample in the current epoch. This loss function was optimized using Stochastic Gradient Descent (SGD) \cite{sgd}.

\begin{algorithm}[h]
\SetAlgoLined
\KwIn{Sample image}
\KwOut{Class of the given image}
\For{$epoch \leftarrow 0$ \KwTo $E$} {
    \uIf{$epoch < E^{\alpha}_{i}$}{
    $\alpha \leftarrow \alpha_{i}$
    } \uElseIf {$epoch < E^{\alpha}_{f}$} {
    $\alpha \leftarrow \frac{\alpha_{f}-\alpha_{i}} {E^{\alpha}_{f}-E^{\alpha}_{i}} * (epoch-E^{\alpha}_{i})+\alpha_{i}$
    }
    \Else {
    $\alpha \leftarrow \alpha_{f}$
    }
    Run the model on train set \\
    $loss \leftarrow BCE(l, \hat{l}) + \alpha * BCE(u_{epoch}, u_{epoch-1})$\\
    Generate the pseudo labels for unlabeled data\\
    Evaluate the model on validation set
}
\caption{Semi-supervised classification train loop}
\label{alg:semi_cls}
\end{algorithm}

\subsection{Experiments}
We used ResNet18 as it is computationally efficient. We experimented with Adam \cite{adam} optimizer and SGD. Optimizing using SGD was much more stable and the optimizer was less susceptible to overshooting. Different values of $\alpha$ were experimented with and it was found that a slow and gradual increase in alpha was better for training the model. Our best performing model uses $\alpha_{i} = 0$ and $\alpha_{f} = 1$. The value of $\alpha$ increases from epoch $E^{\alpha}_{i} = 10$ to $E^{\alpha}_{f} = 135$. The model was trained on batch size of 64.

\begin{table}[!htb]
\begin{tabular*}{\linewidth}{l c c c}
\toprule
Model          & Training          & Test          & \#params \\
               & Accuracy          & Accuracy      & \\\midrule
InceptionNetv3 & 99.03\%           & 84.38\%       & 23.8M\\
ResNet50       & 97.37\%           & 93.69\%       & 25.6M\\
Xception       & \textbf{99.84}\%           & 90.62\%       & 22.9M\\
\textbf{ResNet18 (our)} & 96.69\%  & \textbf{96.70\%} & \textbf{11.6M}\\
\bottomrule
\end{tabular*}
\caption{Classification models comparison}
\label{table:cls_res}
\end{table}

\begin{table*}[!tbh]
\centering
\addtolength{\tabcolsep}{-3.15pt}
\begin{tabular}{p{2.65cm}*{10}c|c}
\toprule
Method & Background  & Building  & Building Non & Road & Road Non & Water & Tree & Vehicle & Pool & Grass & \textbf{mIoU} \\
       &          & Flooded   & Flooded  & Flooded &  Flooded &       &      &         &      &       & \\
\midrule
UNet &0.&0.&0.34&0.&0.45&0.49&0.47&0.&0.&0.64&0.239\\
PSPNet &0.04&0.45&0.66&0.32&0.73&0.61&0.71&0.14&0.18&0.82&0.4665\\
DeepLabV3+ &0.16&\textbf{0.49}&\textbf{0.69}&0.45&\textbf{0.76}&\textbf{0.72}&\textbf{0.76}&0.14&\textbf{0.18}&\textbf{0.85}&0.5204\\
\textbf{DeepLabV3+ (pseudo-labels)}&\textbf{0.17}&0.48&\textbf{0.69}&\textbf{0.48}&0.75&\textbf{0.72}&\textbf{0.76}&\textbf{0.15}&\textbf{0.18}&\textbf{0.85}&\textbf{0.5223}\\
\bottomrule
\addtolength{\tabcolsep}{3.15pt}
\end{tabular}
\caption{Classwise segmentation results on FloodNet testing set}
\label{table:seg_res}
\end{table*}

\begin{figure*}[!tbh]
    \centering
    \includegraphics[width=\textwidth]{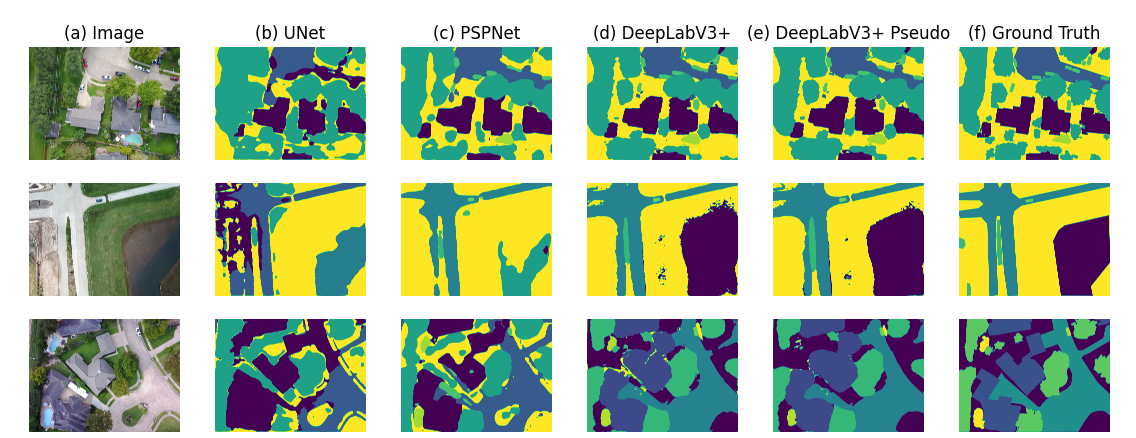}
    \includegraphics[width=15cm]{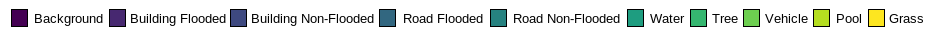}
    \caption{Visual comparison on FloodNet dataset for semantic segmentation}
    \label{fig:seg_ex}
\end{figure*}

\subsection{Results}
Our system performed significantly better than all the classification baseline results mentioned in the FloodNet paper while having a considerably smaller architecture (half the number of parameters) as shown in Table \ref{table:cls_res}. Our best model achieves \textbf{98.10\% F1} and \textbf{96.70\% accuracy} on the test set.

\section{Segmentation}
\label{section:segmentation}
In this section, we detail our approach for training a model which generates multi-class segmentation masks for given images. The semantic labels for the task is a 10 pixel-level class segmentation mask consisting of Background, Building-flooded, Building non-flooded, Road-flooded, Road non-flooded, Water, Tree, Vehicle, Pool and Grass classes. They are mapped from 0 to 9 respectively.

\subsection{Data and Preprocessing}
To expedite the process of feature extraction for the deep learning model, we apply bilateral filter to the image, followed by two iterations of dilation and one iteration of erosion. For image augmentation we perform shuffling, rotation, scaling, shifting and brightness contrast. The images and masks are resized to $512\times512$ dimensions while training since high-resolution images preserve useful information.

\subsection{Methodology}
The dataset contains labeled masks of dimension $3000\times4000\times3$ with pixel values ranging from 0 to 9, each denoting a particular semantic label.  These are one-hot encoded to generate labels with 10 channels, where $i^{th}$ channel contains information about $i^{th}$ class. 

We experiment with various encoder-decoder and pyramid pooling based architectures to train our model, the details of which are mentioned in Section \ref{section:seg_experiments}. The loss function used is a weighted combination of Binary Cross-Entropy loss (BCE) and Dice loss as it provides visually cleaner results.

We apply a semi-supervised approach and generate pseudo masks for the unlabeled images. While training the model for $E$ epochs, the labeled samples were used for $E^{\alpha}_{i}$ epochs where Adam is used as an optimizer. After that pseudo masks were used to further train the model as described in Algorithm \ref{alg:semi_cls}. $\alpha$ has an initial value of $\alpha_{i}$ that increases upto $\alpha_{f}$ from epoch $E^{\alpha}_{i}$ to $E^{\alpha}_{f}$. SGD optimizer with 0.01 LR is used when pseudo masks are introduced to the model.

\subsection{Experiments}
\label{section:seg_experiments}
We apply three state-of-the-art semantic segmentation models on the FloodNet dataset. We adopt one encoder-decoder based network named UNet, one pyramid pooling module based network PSPNet and the last network model DeepLabV3+ employs both encoder-decoder and pyramid pooling based module. We train all of them in a supervised fashion.
For UNet, PSPNet and DeepLabV3+ the backbones used were ResNet34 , ResNet101  and EfficientNet-B3 respectively.

For UNet the learning rate was $0.01$ with step LR scheduler set at intervals [10,30,50] and decay factor $\gamma$ set to $0.1$. For PSPNet the learning rate was $0.001$ without any LR decay. For DeepLabV3+ the learning rate was 0.001 with step LR scheduler set at intervals [7,20] and $\gamma$ set to $0.1$. 

Adam optimizer and batch size of 24 was used for all the models with Mean Intersection over Union (MIoU) as the evaluation metric. We observed the best results when we weighed the BCE loss and Dice loss equally.

Once we recognized the best performing model on the task, we trained a DeepLabV3+ model with EfficientNet-B3 as the backbone and used SGD optimizer instead of Adam optimizer in a semi-supervised fashion. Due to computation and memory constraints we randomly sampled unlabeled data with the ratio of $1:10$ for generating the pseudo masks.

\subsection{Results}

Table \ref{table:seg_res} showcases the comparison of the best models we achieved for each of the 3 architectures. The best test set result  was achieved by a DeepLabV3+ architecture with EfficientNet-B3 backbone. A few example images of the predictions of the model against the ground truth are provided in Figure \ref{fig:seg_ex}. It is evident from the table that small objects like vehicles and pools are the most difficult tasks for our models.

\section{Conclusion} 
\label{section:conclusion}
In this work, we have explored methods to approach semi-supervised classification and segmentation along with handling the class imbalance problem on high-resolution images. We have conducted a range of experiments to obtain the best possible technique and models to optimize for the tasks. 

Our classification framework achieves laudable results with just 398 labeled images and also utilizes the entirety of the unlabeled data. Our segmentation framework shows an increase of 0.19\% on using the unlabeled data as pseudo labels. This provides a wide scope of improvement. The amount of unlabeled data is approximately three times the amount of labeled data which if employed efficiently can produce superior results.

We foresee multiple opportunities for future research. Training the model in an unsupervised fashion and fine-tuning it with the labeled data followed by distillation as presented in SimCLRv2 \cite{simclrv2} is a very promising method. Training with a contrastive loss \cite{contrastive_loss} has been incremental at times. With the emergence of Visual Transformers, self-supervised Vision Transformers \cite{dino} could also be explored for this task.

\section{Acknowledgement}
We would like to thank Research Society Manipal for their valuable inputs and research guidance.

{\small
\bibliographystyle{ieee_fullname}
\bibliography{FLOODNET}

\begin{thebibliography}{10}\itemsep=-1pt

\bibitem{dino}
Mathilde Caron, Hugo Touvron, Ishan Misra, Herv\'e J\'egou, Julien Mairal,
  Piotr Bojanowski, and Armand Joulin.
\newblock Emerging properties in self-supervised vision transformers.
\newblock {\em arXiv preprint arXiv:2104.14294}, 2021.

\bibitem{entropy_reg}
Olivier Chapelle, Bernhard Scholkopf, and Alexander Zien, editors.
\newblock {\em Semi-Supervised Learning}.
\newblock The {MIT} Press, Sept. 2006.

\bibitem{deeplabv3+}
Liang-Chieh Chen, Yukun Zhu, George Papandreou, Florian Schroff, and Hartwig
  Adam.
\newblock Encoder-decoder with atrous separable convolution for semantic image
  segmentation.
\newblock In {\em Proceedings of the European Conference on Computer Vision
  (ECCV)}, September 2018.

\bibitem{simclrv2}
Ting Chen, Simon Kornblith, Kevin Swersky, Mohammad Norouzi, and Geoffrey~E.
  Hinton.
\newblock Big self-supervised models are strong semi-supervised learners.
\newblock {\em CoRR}, abs/2006.10029, 2020.

\bibitem{imagenet}
Jia Deng, Wei Dong, Richard Socher, Li-Jia Li, Kai Li, and Li Fei-Fei.
\newblock Imagenet: A large-scale hierarchical image database.
\newblock In {\em 2009 IEEE Conference on Computer Vision and Pattern
  Recognition}, pages 248--255, 2009.

\bibitem{resnet}
Kaiming He, Xiangyu Zhang, Shaoqing Ren, and Jian Sun.
\newblock Deep residual learning for image recognition.
\newblock In {\em 2016 IEEE Conference on Computer Vision and Pattern
  Recognition (CVPR)}, pages 770--778, 2016.

\bibitem{contrastive_loss}
Prannay Khosla, Piotr Teterwak, Chen Wang, Aaron Sarna, Yonglong Tian, Phillip
  Isola, Aaron Maschinot, Ce Liu, and Dilip Krishnan.
\newblock Supervised contrastive learning.
\newblock {\em CoRR}, abs/2004.11362, 2020.

\bibitem{adam}
Diederik~P. Kingma and Jimmy Ba.
\newblock Adam: {A} method for stochastic optimization.
\newblock In Yoshua Bengio and Yann LeCun, editors, {\em 3rd International
  Conference on Learning Representations, {ICLR} 2015, San Diego, CA, USA, May
  7-9, 2015, Conference Track Proceedings}, 2015.

\bibitem{sudo_lbl}
Dong-Hyun Lee.
\newblock Pseudo-label : The simple and efficient semi-supervised learning
  method for deep neural networks.
\newblock {\em ICML 2013 Workshop : Challenges in Representation Learning
  (WREPL)}, 07 2013.

\bibitem{r2021bert}
Sidharth R, Abhiraj Tiwari, Parthivi Choubey, Saisha Kashyap, Sahil Khose,
  Kumud Lakara, Nishesh Singh, and Ujjwal Verma.
\newblock Bert based transformers lead the way in extraction of health
  information from social media, 2021.

\bibitem{floodnet}
Maryam Rahnemoonfar, Tashnim Chowdhury, Argho Sarkar, Debvrat Varshney, Masoud
  Yari, and Robin~R. Murphy.
\newblock Floodnet: {A} high resolution aerial imagery dataset for post flood
  scene understanding.
\newblock {\em CoRR}, abs/2012.02951, 2020.

\bibitem{sgd}
Herbert Robbins and Sutton Monro.
\newblock {A Stochastic Approximation Method}.
\newblock {\em The Annals of Mathematical Statistics}, 22(3):400 -- 407, 1951.

\bibitem{unet}
Olaf Ronneberger, Philipp Fischer, and Thomas Brox.
\newblock U-net: Convolutional networks for biomedical image segmentation.
\newblock In Nassir Navab, Joachim Hornegger, William~M. Wells, and
  Alejandro~F. Frangi, editors, {\em Medical Image Computing and
  Computer-Assisted Intervention -- MICCAI 2015}, pages 234--241, Cham, 2015.
  Springer International Publishing.

\bibitem{efficientnet}
Mingxing Tan and Quoc Le.
\newblock {E}fficient{N}et: Rethinking model scaling for convolutional neural
  networks.
\newblock In Kamalika Chaudhuri and Ruslan Salakhutdinov, editors, {\em
  Proceedings of the 36th International Conference on Machine Learning},
  volume~97 of {\em Proceedings of Machine Learning Research}, pages
  6105--6114. PMLR, 09--15 Jun 2019.

\bibitem{pspnet}
Hengshuang Zhao, Jianping Shi, Xiaojuan Qi, Xiaogang Wang, and Jiaya Jia.
\newblock Pyramid scene parsing network.
\newblock In {\em Proceedings of the IEEE Conference on Computer Vision and
  Pattern Recognition (CVPR)}, July 2017.

\end{thebibliography}
}
\end{document}